\documentclass[journal]{IEEEtran}
\usepackage[T1]{fontenc}% optional T1 font encoding
\usepackage{graphicx}
\usepackage{times}
\usepackage{helvet}
\usepackage{courier}
\usepackage{amsmath}
\usepackage{algorithm}
\usepackage{algorithmic}
\usepackage{csquotes} 
\usepackage{color}
\usepackage{paralist}
\usepackage{amssymb}
\usepackage{indentfirst}
\usepackage{subfigure}
\usepackage{float}
\usepackage{multirow}
\usepackage{cite}
\usepackage{mathrsfs}

\usepackage{mathrsfs} 
\usepackage{amsfonts}
\usepackage{todonotes}
\usepackage{pgfplots} %引用包
\usepackage{epstopdf}
\usepackage{epsfig}
\pgfplotsset{compat=newest}
\usepackage{color}
\definecolor{forestgreen}{RGB}{0,139,69}

\usepackage{xcolor}
\definecolor{citecolor}{HTML}{0071bc}
\usepackage[colorlinks, linkcolor=red,  anchorcolor=blue, citecolor=citecolor]{hyperref} 

\usepackage{xcolor}
\definecolor{SeaGreen4}{RGB}{0,205,102} 
\definecolor{SlateBlue}{RGB}{106,90,205} 
\definecolor{DarkRed}{RGB}{178,34,34} 
	
\usepackage[switch]{lineno}
\usepackage{textcomp,booktabs}
\usepackage{amssymb}% http://ctan.org/pkg/amssymb
\usepackage{pifont}% http://ctan.org/pkg/pifont

\usepackage{makecell}

\usepackage{colortbl}
\definecolor{mygray}{gray}{.9}
\definecolor{mypink}{rgb}{.99,.91,.95}
\definecolor{mycyan}{cmyk}{.3,0,0,0}

\begin{document}

\title{ 
    EPRBench: A High-Quality Benchmark Dataset for Event Stream Based Visual Place Recognition 
}

\author{   
    Xiao Wang, \textit{Member, IEEE}, Xingxing Xiong, Jinfeng Gao, Xufeng Lou, Bo Jiang,  
    Sibao Chen, \\ Yaowei Wang, \textit{Member, IEEE}, Yonghong Tian, \textit{Fellow, IEEE} \\ 
\thanks{ $\bullet$ Xiao Wang, Xingxing Xiong, Jinfeng Gao, Xufeng Lou, Bo Jiang, Sibao Chen are with School of Computer Science and Technology, Anhui University, Hefei, China. (email: \{xiaowang, jiangbo, sbchen\}@ahu.edu.cn, \{e24201089, e24201028\}@stu.ahu.edu.cn, gaoelieen750@gmail.com)}    
\thanks{ $\bullet$ Yaowei Wang is with Harbin Institute of Technology, Shenzhen, China; Peng Cheng Laboratory, Shenzhen, China. (email: wangyw@pcl.ac.cn)}   
\thanks{ $\bullet$ Yonghong Tian is with Peng Cheng Laboratory, Shenzhen, China; School of Computer Science, Peking University, China; Shenzhen Graduate School, Peking University, China. (email: yhtian@pku.edu.cn)}   
\thanks{ $\bullet$ Corresponding author: Bo Jiang}  
}

\markboth{ IEEE Transactions on XXX, 2026 } 
{Shell \MakeLowercase{\textit{et al.}}: Bare Demo of IEEEtran.cls for IEEE Journals}

% make the title area
\maketitle

% As a general rule, do not put math, special symbols or citations in the abstract or keywords.
\begin{abstract}
Event stream-based Visual Place Recognition (VPR) is an emerging research direction that offers a compelling solution to the instability of conventional visible-light cameras under challenging conditions such as low illumination, overexposure, and high-speed motion. Recognizing the current scarcity of dedicated datasets in this domain, we introduce EPRBench, a high-quality benchmark specifically designed for event stream–based VPR. EPRBench comprises 10K event sequences and 65K event frames, collected using both handheld and vehicle-mounted setups to comprehensively capture real-world challenges across diverse viewpoints, weather conditions, and lighting scenarios. To support
semantic-aware and language-integrated VPR research, we provide LLM-generated scene descriptions, subsequently refined through human annotation, establishing a solid foundation for integrating LLMs into event-based perception pipelines. To facilitate systematic evaluation, we implement and benchmark 15 state-of-the-art VPR algorithms on EPRBench, offering a strong baseline for future algorithmic comparisons. Furthermore, we propose a novel multi-modal fusion paradigm for VPR: leveraging LLMs to generate textual scene descriptions from raw event streams, which then guide spatially attentive token selection, cross-modal feature fusion, and multi-scale representation learning. This framework not only achieves highly accurate place recognition but also produces interpretable reasoning processes alongside its predictions, significantly enhancing model transparency and explainability. 
%%%% 
The dataset and source code will be released on \url{https://github.com/Event-AHU/Neuromorphic_ReID}  
\end{abstract}

\begin{IEEEkeywords}
Visual Place Recognition; Event Stream; Benchmark Dataset; Chain-Of-Thought Reasoning; Large Language Model 
\end{IEEEkeywords}

\IEEEpeerreviewmaketitle

\section{Introduction}

\IEEEPARstart{V}{isual} Place Recognition (VPR)~\cite{VPRSurvey2025},~\cite{wang2022hybrid},~\cite{jin2025edtformer} is a computer vision task that aims to determine whether an agent is at a previously visited location solely based on current visual input, fundamentally framed as a large-scale image retrieval problem. As a cornerstone for long-term, GPS-denied localization in mobile and autonomous systems, VPR finds broad applications in loop closure detection for robotic SLAM, high-precision localization in autonomous driving, UAV navigation, and augmented reality. However, real-world challenges, particularly ``perceptual aliasing” caused by variations in illumination, season, weather, and dynamic occlusions, pose significant robustness hurdles. In addition, advancing VPR requires addressing critical research directions: improving generalization and interpretability in unseen environments (e.g., by integrating semantic priors from large language models and chain-of-thought reasoning), while simultaneously ensuring computational efficiency and scalability for large-scale, long-term deployment.

\begin{figure*}
    \centering
    \includegraphics[width=1\linewidth]{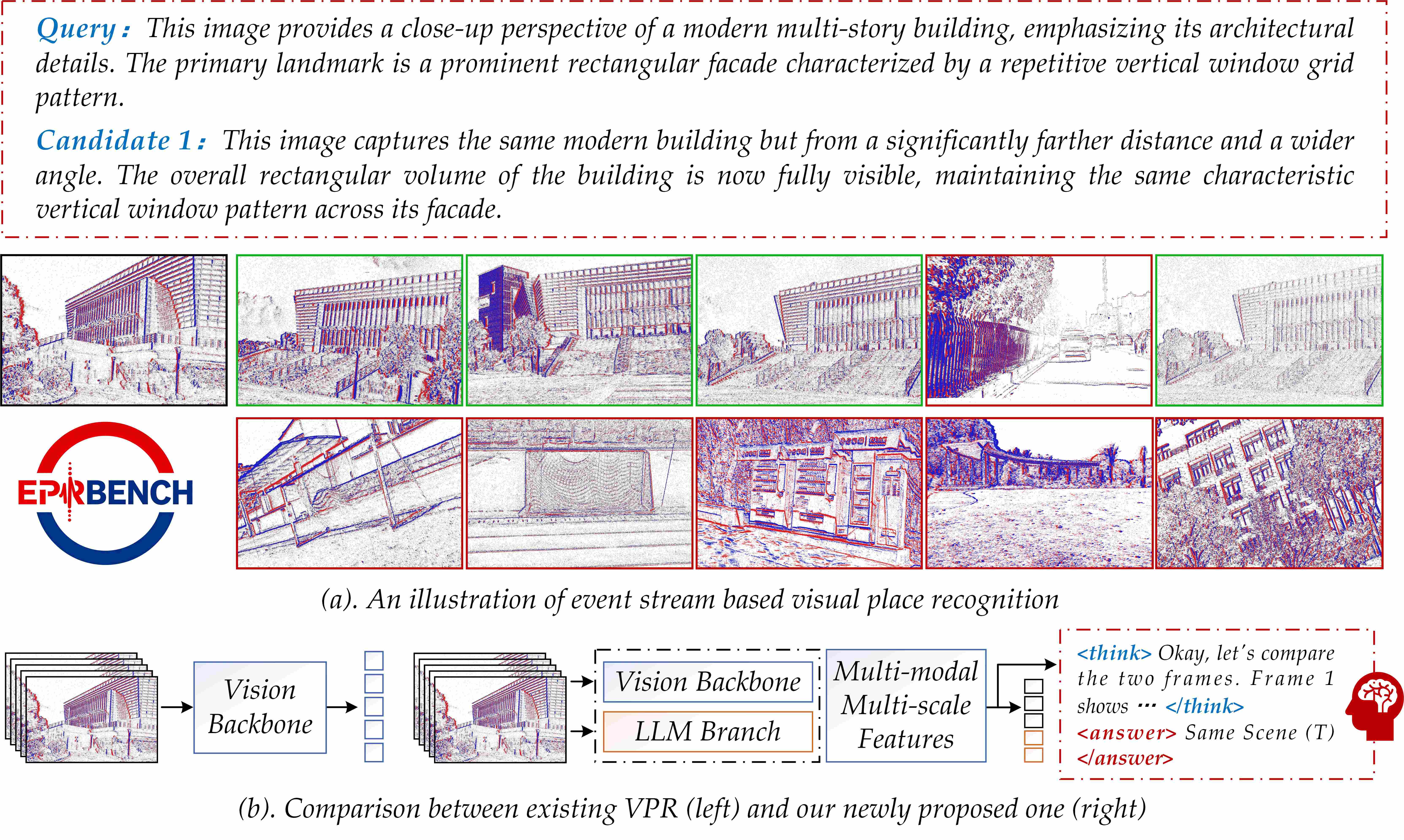}
    \caption{(a). An illustration of event stream-based VPR; (b). Comparison between existing VPR frameworks and ours.} 
    \label{fig:firstIMG}
\end{figure*}

%% event based vision 
Despite significant breakthroughs enabled by deep learning and large foundation models~\cite{oquab2023dinov2}, overall performance remains constrained by the widespread reliance on RGB frame-based cameras. Specifically, RGB cameras are highly sensitive to lighting conditions and perform poorly in scenarios involving overexposure, low illumination, or those requiring high dynamic range (HDR) perception. Moreover, conventional RGB cameras are limited by their frame rate (typically around 30 FPS), making them prone to motion blur under high-speed movement. While using high frame rate cameras could mitigate this issue, it would impose significantly higher demands on data transmission bandwidth and computational processing challenges that hinder practical deployment in real-world applications.

To address these issues, the recently developed event cameras are drawing more and more attention in computer vision~\cite{wang2025evraindrophypergraphguidedcompletioneffective, wang2025reasoningtrackchainofthoughtreasoninglongterm}, such as object detection and tracking~\cite{wang2025longtermvisualobjecttracking}, human activity recognition~\cite{wang2024eventstreambasedhuman}, attribute recognition~\cite{wang2025pedestrianattributerecognitionhierarchical}, person re-identification~\cite{wang2025personreidentificationmeetsevent}, and video captioning~\cite{wang2025estrcotexplainableaccurateevent}. The widespread attention stems from the fact that this sensor is a novel bio-inspired camera capable of asynchronously capturing changes in pixel brightness, thereby directly outputting information about actual motion in the scene, akin to optical flow derived straight from the sensor itself.  Event cameras typically offer an exceptionally wide dynamic range of 120–140 dB, enabling superior perception in both high-exposure and low-light conditions. Their spatial sparsity combined with temporal density allows them to effectively capture high-speed motion; for instance, a rapidly moving drone can acquire clearer, more detailed images of the ground below.  Moreover, event cameras have demonstrated distinct advantages in terms of low power consumption and ultra-low latency. Some researchers also incorporate such sensors into the VPR community, e.g., Lee et al.~\cite{EventVLAD2021} achieve robust VPR in the low-illumination conditions using the textural information from the event cameras. Tobias et al.~\cite{ensemble-event-vpr} propose the first event-based dataset for large-scale visual place recognition, termed Brisbane-Event-VPR. Building upon these valuable datasets and advances, Wang et al.~\cite{wang2025focus} propose the Focus on Local (FOL) approach, which mines reliable discriminative local regions to improve VPR retrieval and re-ranking performance. Ali-Bey et al.~\cite{Ali-bey_2024_CVPR} propose Bag-of-Queries method, which learns a set of global learnable queries to probe input features via cross-attention for consistent information aggregation and achieve superior performance.

%% issues of current event-based vpr 
The aforementioned works have clearly laid a solid foundation, both in terms of data and algorithms, for the task of event-based Visual Place Recognition (VPR), and have demonstrated the research value of this direction. However, progress in this field remains constrained by several key challenges: 
\textbf{1) Limited Availability and Scale of Datasets:} To date, only three event-based VPR datasets are publicly available in the academic community: Brisbane-Event-VPR~\cite{ensemble-event-vpr} (introduced in 2020), NeuroGPR~\cite{neurogpr} (2023), and NYC-Event-VPR~\cite{pan2025nyc} (2024). As summarized in Table~\ref{tab:dataset_comparison}, these datasets are notably limited in both scale and diversity, falling far short of the growing demand for large-scale, high-quality VPR benchmarks.
\textbf{2) Insufficient Semantic Understanding:} Current approaches typically focus on designing powerful neural networks to extract effective feature representations from input images or event streams. This paradigm, centered on representation learning, reflects the dominant methodology of the first half of the third wave of AI, yet it lacks deep exploitation of semantic cues essential for robust place recognition. 
\textbf{3) Poor Model Interpretability:} Mainstream deep learning–based VPR models often rely on activation heatmaps (e.g., feature map visualizations) to highlight regions of interest, offering limited visual justification for their place matching decisions. However, such post-hoc visual explanations remain far from achieving truly intelligent, human-aligned interpretability.

%% benchmark datasets we proposed in this paper 
In this paper, we propose a new large-scale, high-definition ($1280 \times 720$) event stream based VPR benchmark dataset, termed EPRBench. In more detail, our dataset contains 10K event sequences, which involve 1K urban scenes and 65K event frames. The data were primarily collected using two methods: handheld and vehicle-mounted setups, covering a variety of viewpoints, weather conditions, and lighting scenarios. To better support the interpretability of VPR algorithms in the era of large models, this paper also leverages large language models (i.e., DouBao~\cite{doubao2026}) to generate textual descriptions for the provided data and performs manual verification to ensure the accuracy of these semantic descriptions. A comparison between the proposed dataset and existing ones can be found in Table~\ref{tab:dataset_comparison}. In addition, we also retrain and report 15 state-of-the-art VPR algorithms based on our benchmark dataset, including the CNN-based, Transformer-based, and re-ranking based VPR models.

%% the baselines we propose 

Based on the VPR benchmark dataset proposed in this paper, we also introduce a novel VPR paradigm that leverages large language models to extract semantic descriptions from visual inputs, thereby transforming the traditional purely vision-based perception task into a vision-language fusion problem, as shown in Fig.~\ref{fig:firstIMG}. Specifically, Fig.~\ref{fig:firstIMG} (b) contrasts this with existing frameworks: while traditional methods rely solely on opaque visual embeddings, our approach integrates an LLM branch to generate explicit semantic descriptions for more robust matching. Specifically, we adopt DINO V2 and the CLIP model to extract the event and language features. We then compute the cross-modal correlations between semantic and visual tokens, use these correlations to weight the word tokens, and fuse them with the top-k selected visual tokens. Meanwhile, the unselected visual tokens are repurposed to reshape a global feature map representation. Then, multi-scale features are extracted via average pooling as the final semantic enhanced vision features. Note that we also introduce an LLM decoder to generate the chain-of-thought reasoning process to further improve the model interpretability.

%% contributions 

To sum up, we conclude the main contributions of this paper into the following three aspects: 

1) We propose a large-scale and new high-definition benchmark dataset for the event stream based visual place recognition, termed EPRBench. It contains 13,109 video samples and was collected under various weather and light conditions.  

2) We retrain and report the performance of existing 15 open-sourced VPR models based on our EPRBench, which builds a solid foundation for the research of event camera-based place recognition. 

3) We propose a simple yet effective baseline for event-based visual place recognition (VPR) that leverages semantic cues derived from LLMs, termed SG-VPR. By grounding visual features in LLM-generated semantic descriptions, our approach effectively captures spatially meaningful semantics. This is further enhanced through multi-modal fusion and multi-scale feature integration. Moreover, we incorporate chain-of-thought reasoning to improve model interpretability further.

%% organization of this paper 
\textit{The rest of this paper is organized as follows:} 
In section~\ref{sec::relatedWorks}, we review the existing works related to this paper with a focus on visual place recognition and event-based vision. Then, we dive into the details of the EPRBench benchmark dataset in section~\ref{sec::EPRBench}. We describe the baseline approach proposed in this paper in section~\ref{sec::method} and report the experimental results in section~\ref{sec::experiments}. Finally, we conclude this paper and propose possible research directions in section~\ref{sec::conclusion}.

\begin{table*}[htbp]
\centering
\caption{Comparison of representative event-based datasets for visual place recognition. $\#$ denotes the number of the corresponding item. CL, RA, DT, NT represent Clear sky, Rainy day, Day time, and Night time, respectively. }
\label{tab:dataset_comparison}
\resizebox{\textwidth}{!}{
\begin{tabular}{l|c|l|c|c|c|c|cc|cc}
\toprule
\multirow{2}{*}{\textbf{Dataset}} & \multirow{2}{*}{\textbf{Year}} & \multirow{2}{*}{\textbf{Sensor}} & \multirow{2}{*}{\textbf{Resolution}} & \multirow{2}{*}{\textbf{$\#$Scene}} & \multirow{2}{*}{\textbf{Real}}& \multirow{2}{*}{\textbf{Text}}  
& \multicolumn{2}{c|}{\textbf{Weather}} 
& \multicolumn{2}{c}{\textbf{Lighting}} \\
& & & & & &
& CL & RA 
& DT & NT \\
\midrule
DDD17~\cite{binas2017ddd17} 
& 2017 & DAVIS 346 & 346$\times$260 & 40 & $\checkmark$  & $\times$  & $\checkmark$ & $\times$ & $\checkmark$ & $\times$ \\
DDD20~\cite{hu2020ddd20} 
& 2020 & DAVIS 346 & 346$\times$260 & 215 & $\checkmark$  & $\times$ & $\checkmark$ & $\checkmark$ & $\checkmark$ & $\checkmark$ \\
Brisbane-Event-VPR~\cite{ensemble-event-vpr} 
& 2020 & DAVIS 346 & 346$\times$240 & 6  & $\checkmark$  & $\times$ & $\checkmark$ & $\checkmark$ & $\checkmark$ & $\checkmark$ \\
NeuroGPR~\cite{neurogpr} 
& 2023 & DAVIS 346 & 346$\times$240 & 12  & $\checkmark$  & $\times$ & $\checkmark$ & $\times$ & $\checkmark$ & $\times$ \\
NYC-Event-VPR~\cite{pan2025nyc} 
& 2024 & Prophesee EVK4-HD & 1280$\times$720 & 16  & $\checkmark$  & $\times$ & $\checkmark$ & $\checkmark$ & $\checkmark$ & $\checkmark$ \\
\midrule
EPRBench (Ours) & 2026 & Prophesee EVK4-HD & 1280$\times$720 &  1022 & $\checkmark$ & $\checkmark$ & $\checkmark$ & $\checkmark$ & $\checkmark$ & $\checkmark$ \\ 
\bottomrule
\end{tabular}}
\end{table*}

\section{Related Work} \label{sec::relatedWorks}

\subsection{Visual Place Recognition} 

Visual Place Recognition (VPR) is a task that identifies previously visited locations solely based on visual input, aiming at achieving autonomous localization and relocalization through image information. Traditional VPR algorithms primarily consist of three main components: \textit{feature extraction}, \textit{feature aggregation}, and \textit{feature matching}. Recently, feature extraction methods based on deep learning have made significant progress. Using convolutional neural networks (CNNs)~\cite{he2016deep, simonyan2015very, tan2019efficientnet} to extract intermediate layer features, or leveraging the Vision Transformer (ViT)~\cite{dosovitskiy2021an} architecture to enhance modeling capabilities, it becomes more effective to extract both \textit{local features} (e.g., feature points and descriptors) and \textit{global features} (i.e., a single vector representation used for efficient retrieval). These approaches significantly outperform traditional handcrafted feature methods such as SIFT~\cite{lowe2004distinctive} and SURF~\cite{bay2006surf}.

In the feature aggregation stage, early statistical aggregation methods~\cite{sivic2003video, jegou2010aggregating, perronnin2010improving} were manually designed, lacked learning capabilities, and struggled to adapt to complex environmental changes. NetVLAD~\cite{arandjelovic2016netvlad} addresses this by aggregating local features in an end-to-end trainable manner. GeM pooling~\cite{radenovic2018revisiting} enhances robustness by applying power-average pooling to the CNN output feature maps. MixVPR~\cite{ismail2022mixvpr} adopts feature concatenation and compression of multiple convolutional channels as the aggregation method, eliminating the need for local feature point calculations, thus offering faster and more efficient performance. Additionally, SALAD~\cite{mahjoub2023salad} proposes a new approach by formulating feature aggregation as an optimal transport problem, explicitly excluding less informative local features through a ``dustbin'' mechanism to improve tolerance against dynamic backgrounds and occlusions. There are also methods that incorporate attention mechanisms and transformer architectures, such as TransVPR~\cite{hou2022transvpr} and R2Former~\cite{xu2023r2former}, to enhance spatial modeling and context awareness.

Feature matching primarily involves two strategies: \textit{coarse matching} based on global descriptors and \textit{fine matching} based on local features. Global descriptor matching, currently the mainstream due to its efficiency, involves extracting global vectors from each image~\cite{arandjelovic2016netvlad, radenovic2018revisiting, ismail2022mixvpr} and retrieving the most relevant candidate images using vector similarity measures such as L2 norm or cosine similarity. To further improve matching accuracy, many methods incorporate fine matching strategies based on local features. These extract keypoints and local descriptors~\cite{lowe2004distinctive, detone2018superpoint, dusmanu2019d2}, and perform fine-grained comparisons of candidate images using spatial consistency verification techniques such as RANSAC or geometric transformations. Furthermore, advanced techniques such as sequence matching (e.g., SeqSLAM~\cite{milford2012seqslam}, FAB-MAP with Temporal Priors~\cite{cummins2008fab, cummins2008appearance}) match the current image with a continuous sequence of images, leveraging temporal redundancy to improve robustness and accuracy in dynamic environments. Although deep learning-based VPR methods have achieved excellent performance, they still face challenges in extreme scenarios such as strong lighting conditions or high-speed motion, where standard cameras are prone to glare and motion blur. Furthermore, most existing methods are based on static architectures and lack adaptive feature extraction mechanisms, which can limit their effectiveness in rapidly changing or event-rich environments.

\subsection{Event-based Vision}   

% In recent years, researchers have increasingly focused on utilizing event cameras to solve relevant vision problems due to their low latency, high dynamic range, and high temporal resolution. Beyond the emerging exploration in VPR, event cameras have become a research hotspot in computer vision~\cite{wang2025evraindrophypergraphguidedcompletioneffective,wang2025reasoningtrackchainofthoughtreasoninglongterm}, such as object detection and tracking~\cite{wang2025longtermvisualobjecttracking}, 
% Ensemble-Event-VPR~\cite{ensemble-event-vpr} takes different numbers of events, as well as time windows to reconstruct event intensity frames and visual descriptors. Event-VPR~\cite{event-vpr} proposes an ETS voxel grid to represent the event stream, a deep residual network to extract temporal features, and a modified VLAD network for feature aggregation to achieve end-to-end VPR. EFormer-VPR~\cite{eformer-vpr} proposes event clusters and adaptive event windows for the event prepossessing, and uses a fractional mechanism to align features of different modalities to compensate for the lack of single-mode feature representation when fusing frame and event features. FE-Fusion-VPR~\cite{fe-fusion-vpr} employs a multi-scale fusion network to obtain features at three scales, which are then aggregated into three sub-descriptors through a VLAD layer. Finally, a descriptor weight learning network is utilized to derive the final descriptor, etc. 

Distinct from the synchronous integration of traditional shutters, event sensors operate on a bio-inspired, asynchronous paradigm, enabling robust perception in high-dynamic-range and high-speed scenarios~\cite{chen2022ecsnet}. This mechanism has catalyzed significant progress in fundamental vision tasks, often driven by the co-evolution of large-scale benchmarks and novel algorithms~\cite{wang2025mambaevt},~\cite{liu2023voxel}. In the field of object detection, MvHeat-DET~\cite{wang2024EvDET200K} constructs the large-scale EvDET200K benchmark and proposes a Mixture-of-Experts (MoE) heat conduction backbone, which effectively models feature extraction as a thermal diffusion process to balance accuracy and efficiency. Similarly, for long-term visual object tracking, AMTTrack~\cite{wang2025longtermvisualobjecttracking} contributes the FELT dataset and introduces an Associative Memory Transformer incorporating Modern Hopfield Networks, which enables robust cross-modal retrieval and dynamic template updating to handle significant appearance changes. These works demonstrate that establishing high-quality datasets coupled with semantically aware algorithms is pivotal for advancing event-based vision.
In the specific domain of localization, existing VPR approaches generally fall into reconstruction-based and fusion-based categories. Early works like Ensemble-Event-VPR~\cite{ensemble-event-vpr} adopt a reconstruction strategy, converting asynchronous events into intensity frames via variable time windows to leverage established visual descriptors. Conversely, Event-VPR~\cite{event-vpr} pioneers an end-to-end learning framework, utilizing an Event Spike Tensor (EST) voxel grid to structure the raw stream, followed by a deep residual network and a modified VLAD layer for direct feature aggregation.
In their work, most researchers adopt feature fusion methods to obtain fused features of traditional frames and events, thereby achieving superior visual place recognition performance. Different from their approaches, in our work, we propose a novel event-based visual place recognition benchmark dataset.

\section{EPRBench Dataset} \label{sec::EPRBench} 

In this section, we will first introduce the protocols needed to follow when collecting our event videos. Then, we will focus on data collection and statistical analysis. 

\begin{figure*}
\centering
\includegraphics[width=\textwidth]{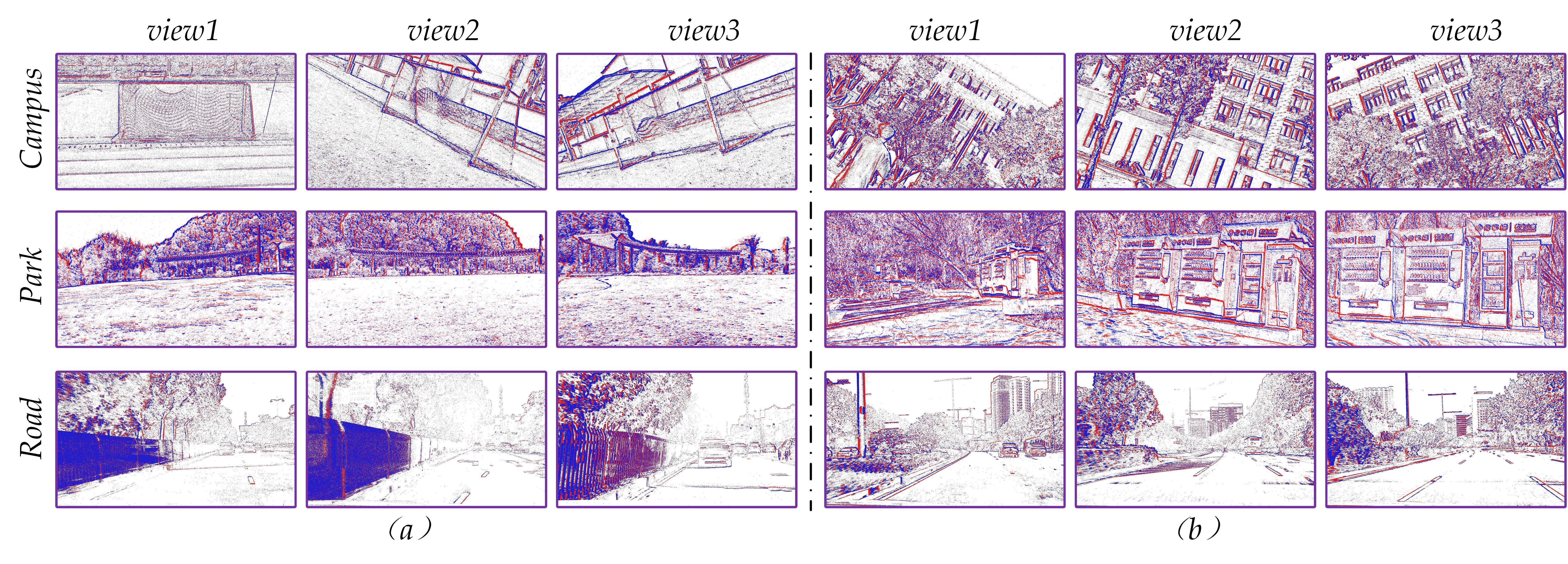}
\caption{Data samples from our newly proposed EPRBench dataset. We visualize two distinct samples (a, b) for each scene category (Campus, Park, Road). For every sample, three images from different viewpoints are presented, demonstrating the significant viewpoint variations contained in our dataset.}
\label{fig:datasets_visualization}
\end{figure*}

\subsection{Protocols}   
In this work, we intend to develop an event-based benchmark dataset for visual scene recognition, providing a data foundation for future research. When constructing this benchmark dataset, we follow the following protocols:
\textit{1) Large-scale:} In the context of current human technological and artificial intelligence development, large-scale datasets have become crucial for the advancement of intelligent human life. To promote the progress of intelligent human life, the dataset we collected includes over 10k event sequences, a large-scale dataset comprising more than 1k urban scenes, and over 65k event frames.
\textit{2) Different Collection Approaches:} We adopt two methods for data collection: the one uses vehicle-mounted event cameras to gather road scenes with wide field views, enabling the capture of rich semantic information in large-scale environments; the other relies on manual data collection to capture fine-grained details in small-scale scenarios, supplementing information unattainable by vehicle-mounted systems to ensure comprehensive coverage of urban environments.
\textit{3) Multi-viewpoint and Multi-condition:} During data acquisition, we captured scene information from multiple viewpoints for each scenario to ensure comprehensive contextual coverage. During the same time, the dataset was collected under various climatic and lighting conditions to introduce challenges, facilitating the construction of a high quality event dataset.

\subsection{Data Collection and Statistical Analysis} 
This EPRBench dataset was acquired with a PROPHESEE EVK4-HD event camera with $1280 \times 720$ resolution. During the data collection period, we always followed the above protocols to ensure that we provide a dataset with event data that contains rich and challenging urban scenarios. We divide the urban scene data according to three main categories: Campus, Park, and Road; where Campus includes scenes inside closed and dense building groups, such as neighborhoods and schools; Park includes non-building open scenes that are not next to the city's main roads, such as parks and attractions; and Road denotes scenes along the streets inside the city as well as scenes near the city's highways. These three types of data almost cover most of the scenes in the city. The EPRBench dataset contains 13,109 event scene samples, where each event sample consists of five event frames, Campus which has 315 scenes containing 2,766 scene samples, Park which has 351 scenes containing 5,176 scene samples, and Road which has 356 scenes containing 5,167 scene samples. The dataset was randomly divided into training, validation, and testing subsets in the ratio 7:1:2, containing 9,220, 1,273, and 2,616 scene samples, respectively.

\subsection{Scenario Text Generation}
\begin{figure*}[!htp]
\centering
\includegraphics[width=\textwidth]{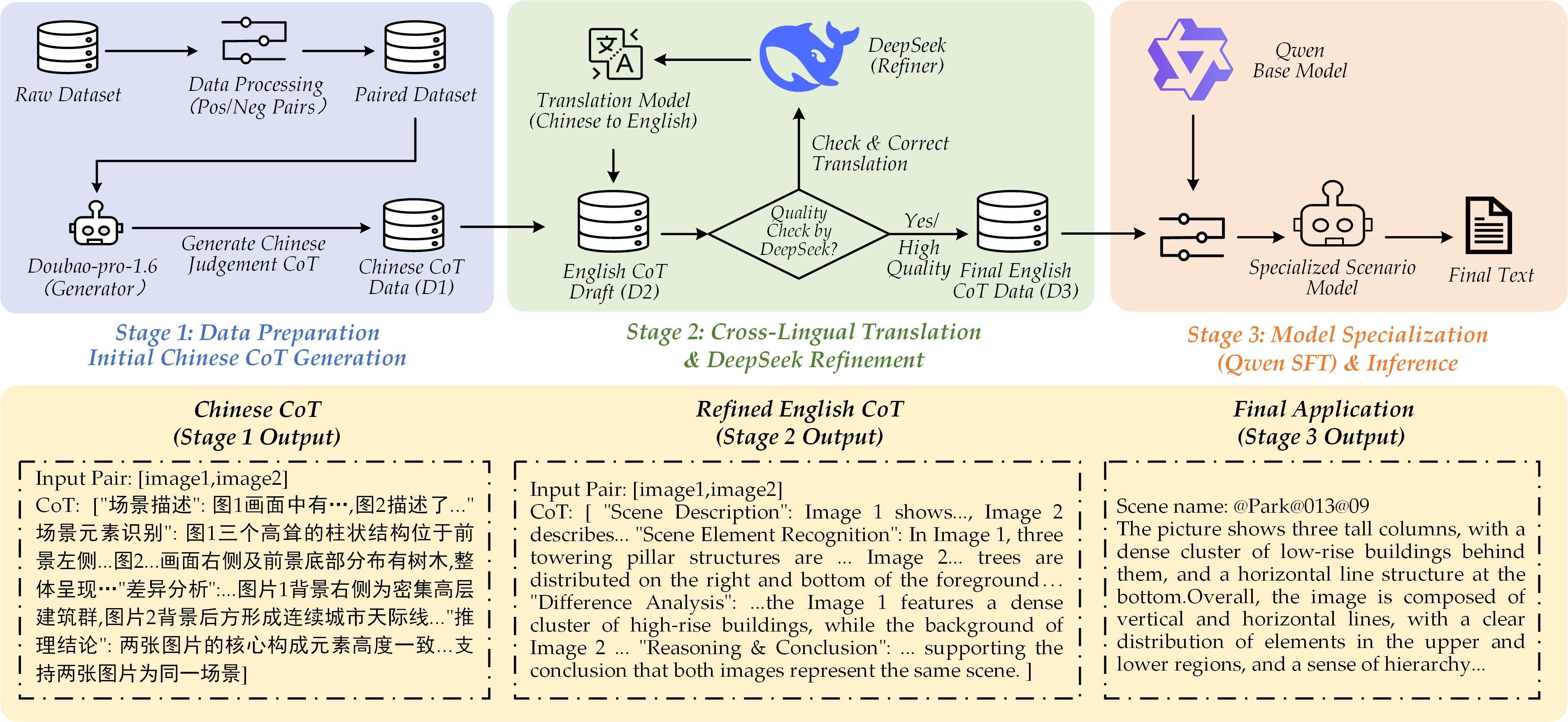}
\caption{The overall pipeline of constructing the specialized scenario description model.We first synthesize Chinese CoT data from paired images, then rigorously translate and refine it into English using DeepSeek to ensure semantic accuracy. Finally, this curated dataset is used to fine-tune the Qwen model, enabling it to perform robust scene reasoning and description.}
\label{fig:text_generation}
\end{figure*}

To augment the visual data with rich semantic information, we propose a multi-stage pipeline to generate specialized scenario descriptions, as illustrated in Fig.~\ref{fig:text_generation}. The process consists of the following three core stages:

\textbf{ Stage 1. Initial Chinese CoT Generation.} 
In this stage, 50,000 image sample pairs are first constructed based on the original event data, with positive and negative samples each accounting for half. Subsequently, text generation is performed on the constructed image sample pairs based on a template prompting strategy, and the Doubao Pro-1.6 model~\footnote{\url{https://www.doubao.com/}} is selected for this generation task. Comparative verification with the output results of other mainstream generation models shows that the text generated by the Doubao Pro-1.6 model can more fully reflect the scene semantic information and has better fine-grained feature capture capability. Finally, the initial Chinese Chain-of-Thought (CoT)~\cite{wei2022chain} dataset is obtained through this stage.

\textbf{ Stage 2. Cross-Lingual Translation and Refinement.} 
The purpose of this stage is to perform cross-lingual conversion and quality optimization on the Chinese CoT dataset obtained in Stage 1, which is specifically divided into two sub-steps: 1) Translation stage: The NLLB-200-3B model~\cite{nllb2024scaling} is adopted as the core translation model to convert the Chinese CoT data into English, obtaining the initial English CoT dataset; 2) Refinement stage: Considering that the initial translation results have quality issues such as expression deviations and grammatical errors, the DeepSeek~\cite{liu2024deepseek} model is further introduced. Through constructing targeted prompt words, the model is guided to verify and correct the translation accuracy and language fluency of the text, and finally a high-quality English scene judgment CoT dataset is obtained.

\textbf{ Stage 3. Model Fine-tuning and Inference.} 
To generate scene description text with rich semantic information, this work constructs a scenario-specific generation model (scene model expert) by fine-tuning a basic vision-language model. In the experiment, Qwen2.5-VL-7B~\cite{bai2025qwen2} is selected as the base model, and the Low-Rank Adaptation (LoRA)~\cite{hu2022lora} fine-tuning strategy (i.e., Supervised Fine-Tuning, SFT) is adopted to fine-tune the parameters of the basic model using the high-quality CoT dataset constructed in Stage 2. The fine-tuned model has efficient scene perception capability and can be used as a scene perception expert model to generate semantically rich description text for the scene images in the dataset.

\subsection{Benchmark Baselines} 

To establish a comprehensive performance reference for event-based visual place recognition, we have selected 15 SOTA or representative VPR algorithms for evaluation on our proposed EPRBench dataset, including:
\textbf{1) CNN-based Global Descriptors:} CosPlace~\cite{cosplace}, MixVPR~\cite{ismail2022mixvpr}, and EigenPlaces~\cite{eigenplaces}. These methods focus on extracting robust global representations from single-mode inputs using convolutional neural networks and have established strong benchmarks in traditional large-scale VPR tasks.
\textbf{2) Transformer-based Local Aggregators:} CricaVPR~\cite{cricavpr}, BoQ~\cite{Ali-bey_2024_CVPR}, and SALAD~\cite{mahjoub2023salad}. These models leverage attention mechanisms or advanced matching-based aggregation to capture fine-grained spatial relationships. Specifically, CricaVPR~\cite{cricavpr} utilizes cross-image correlation, while SALAD~\cite{mahjoub2023salad} employs Sinkhorn-based optimal transport for robust feature alignment.
\textbf{3) Re-ranking based Methods:} R2Former~\cite{xu2023r2former}, SelaVPR~\cite{lu2024towards} and Pair-VPR~\cite{hausler2025pairvpr}. This category includes two-stage pipelines that refine initial retrieval results through sophisticated matching schemes. R2Former~\cite{xu2023r2former} employs a global-to-local reranking scheme with Transformers, and SelaVPR~\cite{lu2024towards} performs semantic label alignment to enforce consistency between queries and candidates, effectively filtering out visually similar but semantically distinct distractors. 

By benchmarking these diverse algorithms, we aim to demonstrate the unique challenges posed by the NYC-Event-VPR and EPRBench benchmarks, especially in scenarios involving extreme motion blur and modality shifts.

\section{Methodology}  \label{sec::method}

\begin{figure*}
\centering
\includegraphics[width=1\linewidth]{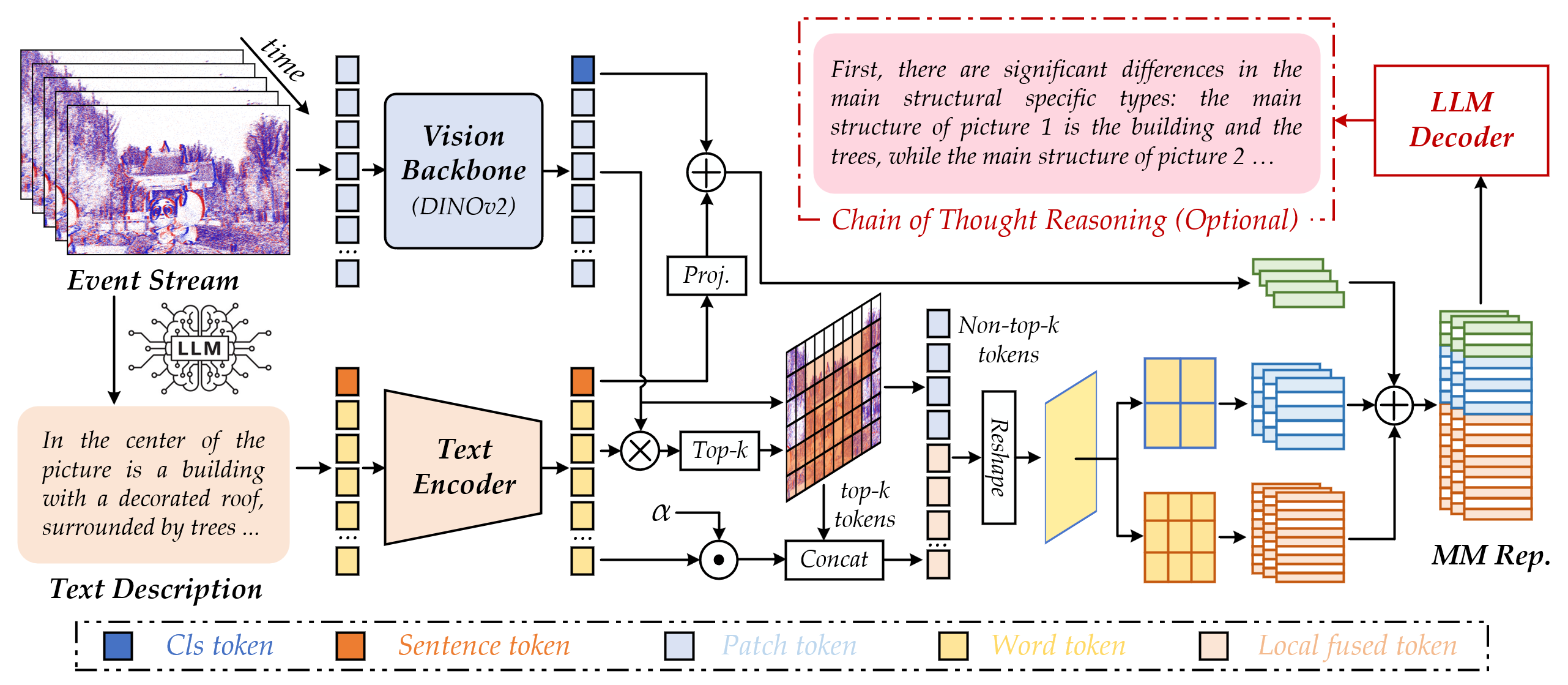}
\caption{An overview of our proposed Semantic Guided VPR framework for event-based place recognition (SG-VPR). The dual-stream system processes event streams and textual descriptions in parallel, adopting a text-guided top-$k$ token selection strategy to extract robust multi-modal representations, with an auxiliary LLM decoder enhancing feature interpretability for retrieval.} 
\label{fig:framework}
\end{figure*}

\subsection{Overview}  
 
The overall architecture of our proposed Semantic-Guided Multi-modal VPR (SG-VPR) framework is illustrated in Fig.~\ref{fig:framework}. It operates as a dual-stream system that processes event streams and textual descriptions in parallel. To extract robust scene representations, we introduce a text-guided token selection strategy. This module leverages the semantic features from the text encoder to evaluate the relevance of visual tokens, applying a top-$k$ strategy to sparsely sample significant regions while discarding task-irrelevant redundancy. The selected visual tokens are then fused with textual features to generate the final multi-modal representation for retrieval. Additionally, an auxiliary LLM decoder is employed to enhance feature interpretability.

\subsection{Network Architecture} 

The overall architecture follows a progressive pipeline structured into four stages: Dual-stream Feature Encoding, Text-Guided Semantic Fusion, Multi-modal Spatial Pyramid Aggregation, and Auxiliary Interpretability Reasoning.

$\bullet$ \textbf{Dual-stream Feature Encoding.~} 
Since the raw event stream $\mathcal{E} = \{e_k\}_{k=1}^{N}$ is asynchronous and spatially sparse, it is incompatible with standard vision backbones. We first employ a temporal aggregation strategy to convert the stream into a synchronous representation. Specifically, distinct events within a fixed time interval $\Delta T$ are accumulated based on their polarity $p_k$ and coordinates $(x_k, y_k)$ to generate a 2D event frame $F \in \mathbb{R}^{H \times W}$. This frame effectively collapses high-temporal motion information into a static spatial structure containing rich edge and texture details. Subsequently, $F$ is fed into the DINOv2~\cite{oquab2023dinov2} encoder, which is robust against illumination changes and requires no human annotations. The encoder patchifies the frame and extracts discriminative features, resulting in local patch embeddings $v_{patch} \in \mathbb{R}^{N \times D}$ and a global representation $v_{cls} \in \mathbb{R}^D$, serving as the visual anchor for place recognition. In parallel, the textual stream processes the semantic description $S = \{w_1, \dots, w_M\}$ (e.g., geometric structures and landmarks) to provide semantic guidance. The raw text is first tokenized and padded to a fixed length $L$ to form an input vector. We then utilize a frozen CLIP Text Encoder~\cite{radford2021learning} to map these discrete tokens into a continuous semantic embedding space. We extract the output from the final transformer layer to obtain the word-level features $t_w \in \mathbb{R}^{L \times D}$ and the sentence-level feature $t_s \in \mathbb{R}^D$, which are crucial for guiding the network to focus on task-relevant regions in the subsequent fusion stage.

$\bullet$ \textbf{Text-Guided Semantic Fusion.~}
To facilitate deep interaction between the visual and textual modalities and ensure robustness against dynamic artifacts, we devise a hierarchical fusion framework. This module synchronizes features at two distinct granularities: Global Context Aggregation for holistic scene definition and Local Sparsification for fine-grained landmark enhancement.

The global fusion module aims to align the high-level semantic representation of the scene with the holistic visual content. Given the global visual class token $v_{cls} \in \mathbb{R}^D$ extracted by the vision backbone and the sentence-level textual embedding $t_s \in \mathbb{R}^D$, we project them into a unified latent space. We then employ a concatenation-based fusion mechanism to synthesize the multi-modal global context:
\begin{equation}
    \mathbf{f}_{global} = \mathcal{F}_{mlp}\left( \text{Concat}(v_{cls}, t_s) \right),
\end{equation}
where $\text{Concat}(\cdot, \cdot)$ denotes the concatenation operation, and $\mathcal{F}_{mlp}(\cdot)$ represents a Multi-Layer Perceptron that models the non-linear interaction between the heterogeneous modalities. This resulting vector $\mathbf{f}_{global}$ encapsulates the overall scene atmosphere, ensuring the network is conditioned on the global semantic prior before processing local details.

While global fusion captures the gist of the scene, event streams often contain significant background clutter and dynamic interference. To address this, the Local Fusion module utilizes fine-grained word-level features $t_w$ to guide the selection and enhancement of informative visual patches $v_{patch}$.
We first evaluate the semantic relevance of each visual patch $v_i$ by  computing its maximum cosine similarity with the word tokens $t_w$, formulated as 
\begin{equation}
    m_i = \max_{j} (\frac{v_i \cdot w_j^T}{\|v_i\| \|w_j\|}).
\end{equation}
Based on these relevance scores, we employ a ratio-based selection strategy to identify the discriminative regions. Specifically, we retain the top $\rho$ percent of patches (where $\rho=0.25$ in our experiments), dynamically determining the selection count $k = \lfloor \rho \cdot N \rfloor$. Let $\mathcal{K}$ denote the indices of these top-$k$ patches, which typically correspond to static landmarks described in the text. Subsequently, we perform semantic injection exclusively on these selected regions to enhance their representation, while keeping unselected background regions unchanged:
\begin{equation}
    \hat{v}_i = 
\begin{cases} 
v_i + \alpha \cdot (v_i \odot w_{idx_i}) & \text{if } i \in \mathcal{K} \\
v_i & \text{otherwise}
\end{cases},
\end{equation}
where $w_{idx_i}$ denotes the word token most similar to patch $v_i$, $\odot$ represents element-wise interaction, and $\alpha$ is a learnable scaling factor. This adaptive fusion design ensures the network focuses its computation on the most semantically significant visual content, effectively discarding task-irrelevant noise.

$\bullet$ \textbf{Multi-modal Spatial Pyramid Aggregation.~} 
To synthesize a unified descriptor, we reshape the patch tokens into a 2D feature map $\mathbf{F}_{map}$ and employ a modified spatial pyramid strategy. Unlike standard approaches, we substitute the global pooling layer with the multi-modal semantic anchor $\mathbf{f}_{global}$, while applying Generalized Mean (GeM) pooling~\cite{radenovic2018fine} to the finer $2\times2$ and $3\times3$ grids to capture local geometry. The final representation $\mathbf{V}_{final}$ is generated via concatenation:
\begin{equation}
\begin{split}
    \mathbf{F}_{final} = \operatorname{Concat}\big( & \mathbf{f}_{global}, \, \operatorname{GeM}_{2\times2}(\mathbf{F}_{map}), \\
    & \operatorname{GeM}_{3\times3}(\mathbf{F}_{map}) \big)
\end{split},
\end{equation}
where $\operatorname{Concat}(\cdot)$ denotes the channel-wise concatenation, and $\operatorname{GeM}_{k\times k}(\cdot)$ represents the aggregated feature vector derived from the $k\times k$ spatial partition. By integrating the global semantic anchor (aligned with textual descriptions) and the local geometric descriptors (derived from event streams), $\mathbf{F}_{final}$ effectively encapsulates both high-level linguistic priors and fine-grained visual structures. This comprehensive multi-modal embedding is directly employed for the final retrieval task, ensuring robustness against extreme environmental variations.

$\bullet$ \textbf{Auxiliary Interpretability Reasoning.~} 
We introduce an optional reasoning branch driven by a trained Scene Description Expert (a specialized LLM decoder). This module translates the aggregated multi-modal representation $\mathbf{F}_{final}$ back into natural language, validating that the learned features preserve high-level semantic fidelity for human-understandable interpretation.

\begin{table*}[htbp]
\centering
\caption{Comparison with other SOTA visual place recognition models.} 
\label{tab:comparison-methods}   
\resizebox{\textwidth}{!}{ 
\begin{tabular}{l|l|ccc|ccc|ccc}
\toprule
\multirow{2}{*}{\textbf{Method}} & \multirow{2}{*}{\textbf{Source}} & \multicolumn{3}{c|}{\textbf{NYC-Event-VPR-Event}} & \multicolumn{3}{c|}{\textbf{NYC-Event-VPR-RGB}} & \multicolumn{3}{c}{\textbf{EPRBench}} \\
& & R@1 & R@5 & R@10 & R@1 & R@5 & R@10 & R@1 & R@5 & R@10 \\
\midrule
01. Deep visual geo-loc~\cite{deepvg} & CVPR'22 & 25.6 & 41.9 & 49.1 & 84.9 & 89.3 & 90.4 & 63.2 & 77.9 & 82.8 \\
02. CosPlace~\cite{cosplace} & CVPR'22 & 32.1 & 48.1 & 55.3 & 85.4 & 88.8 & 89.7 & 73.2 & 88.9 & 92.0 \\
03. MixVPR~\cite{ismail2022mixvpr} & WACV'23 & 45.7 & 63 & 68.9 & 88.3 & 91.0 & 91.5 & 90.2 & 93.7 & 95.6  \\
04. EigenPlaces~\cite{eigenplaces} & ICCV'23 & 28.5 & 43.4 & 51.2 & 81.4 & 91.7 & 93.9 & 81.1 & 86.0 & 89.0 \\
05. R2Former~\cite{xu2023r2former} & CVPR'23 & 61.2 & 75.0 & 78.5 & 86.3 & 88.9 & 89.9 & 80.0 & 94.8 & 96.7 \\
06. CricaVPR~\cite{cricavpr}& CVPR'24 & 30.3 & 48.2 & 55.0 & 88.3 & 92.2 & 92.9 & 92.6 & 95.3 & 96.5 \\
07. SelaVPR~\cite{lu2024towards}& ICLR'24 & 43.3 & 57.6 & 62.9 & 87.8 & 91.3 & 91.7 & 89.9 & 93.6 & 95.2 \\ 
08. EMVP~\cite{qiu2024emvp} & NeurIPS'24 & 45.7 & 64.1 & 70.9 & 85.5 & 90.4 & 91.4 & 88.7 & 93.9 & 95.7 \\
09. SALAD~\cite{mahjoub2023salad}& CVPR'24 & 39.6 & 57.8 & 64.1 & 89.1 & 91.8 & 92.3 & 91.3 & 93.4 & 95.3 \\
10. SuperVLAD~\cite{lu2024supervlad}& NeurIPS'24 & 39.7 & 59.2 & 66.6 & 88.8 & 91.5 & 92.4 & 93.3 & 95.0 & 96.2 \\
11. BoQ~\cite{Ali-bey_2024_CVPR} & CVPR'24 &38.9  &56.7  &63.6  &83.2  &88.1  &89.1  & 93.1 & 94.7 & 95.7 \\
12. ImAge~\cite{ImAge} & NeurIPS'25 &42.4  &59.0  &65.0  &85.7  &90.4  &91.0  & 76.2  & 89.7 & 92.7\\
13. PRGS~\cite{zuo2025prgs} & PR'25 & 43.8 & 62.1 & 69.3 & 82.8 & 88.6 & 89.9 & 87.7  & 93.9 & 96.9\\
14. Pair-VPR~\cite{hausler2025pairvpr} & RA-L'25 & 64.0 & 75.2 & 78.4 & 89.4 & 91.9 & 92.4 & 91.9  & 95.7 & 95.8\\
15. FOL~\cite{wang2025focus} & AAAI'25 & 55.5 & 68.2 & 72.5 & 85.1 & 89.3 & 89.9 & 93.8  & 96.0 & 97.1\\
\midrule
16. SG-VPR (Ours) & - & 57.5 & 74.1 & 78.9 & 86.5 & 90.9 & 92.0 & 94.3 & 96.1 & 97.1 \\
\bottomrule
\end{tabular}} 
\end{table*}

\subsection{Loss Function} 

We train the proposed framework using a joint objective that combines metric learning for retrieval and contrastive learning for cross-modal alignment.

$\bullet$ \textbf{VPR Metric Loss.~} 
To optimize the retrieval performance, we employ the Multi-Similarity (MS) Loss~\cite{wang2019multi}, which adaptively weights hard positive and negative pairs. Let $S_{ik} = \langle \mathbf{f}_i, \mathbf{f}_k \rangle$ be the cosine similarity between descriptors. The loss is defined as:
\begin{equation}
\begin{split}
    \mathcal{L}_{ms} = \frac{1}{B} \sum_{i=1}^{B} \bigg\{ & \frac{1}{\alpha} \log \left[ 1 + \sum_{k \in \mathcal{P}_i} e^{-\alpha (S_{ik} - \lambda)} \right] \\
    & + \frac{1}{\beta} \log \left[ 1 + \sum_{k \in \mathcal{N}_i} e^{\beta (S_{ik} - \lambda)} \right] \bigg\}
    \end{split},
\end{equation}
where $\mathcal{P}_i$ and $\mathcal{N}_i$ denote the positive and negative sample sets for anchor $i$. We set $\alpha=1.0, \beta=50, \lambda=1.0$.

$\bullet$ \textbf{Cross-modal Alignment Loss.~} 
To enforce semantic consistency, we apply a symmetric InfoNCE Loss~\cite{oord2018representation} between the global visual token $v_{cls}$ and global textual token $t_{s}$:
\begin{equation}
\label{equation8}
\begin{split}
    \mathcal{L}_{con} = - \frac{1}{2B} \sum_{i=1}^{B} \bigg( & \log \frac{e^{\langle \mathbf{v}_i, \mathbf{t}_i \rangle / \tau}}{\sum_{j} e^{\langle \mathbf{v}_i, \mathbf{t}_j \rangle / \tau}} \\
    & + \log \frac{e^{\langle \mathbf{t}_i, \mathbf{v}_i \rangle / \tau}}{\sum_{j} e^{\langle \mathbf{t}_i, \mathbf{v}_j \rangle / \tau}} \bigg)
\end{split},
\end{equation}
where $\tau=0.07$ is the temperature parameter. The final loss is a weighted sum:
\begin{equation}
\label{e8}
    \mathcal{L}_{total} = \mathcal{L}_{ms} + \gamma \cdot \mathcal{L}_{con},
\end{equation}
where $\gamma=0.15$ balances the two terms.

\section{Experiments} \label{sec::experiments}

\subsection{Dataset and Evaluation Metric}  

To evaluate the effectiveness of our proposed method, we conducted extensive experiments on two datasets: \textbf{NYC-Event-VPR}~\cite{pan2025nyc} and \textbf{EPRBench} dataset.
Specifically, considering that NYC-Event-VPR contains multi-modal data (both RGB and Event streams), we performed separate experiments on both its RGB and Event modalities to ensure a comprehensive assessment. Specifically, EPRBench is the first event-based VPR dataset to provide rich semantic annotations for each scene, facilitating the cross-modal alignment between sparse event streams and high-level textual descriptions. By incorporating complex environmental noise and drastic appearance changes, EPRBench establishes a rigorous and comprehensive benchmark for evaluating the robustness, interpretability, and viewpoint-invariance of multi-modal localization algorithms in challenging real-world scenarios.

We employ Recall@N as the primary metric to evaluate the performance of our proposed method. For each query, we retrieve the $N$ nearest candidates from the database based on the similarity of their feature descriptors. A retrieval is considered a true positive if at least one of the top-$N$ retrieved samples shares the same location label as the query. The Recall@N score is then calculated as the percentage of successfully identified queries over the entire query set. Following the standard protocol in VPR research, we report the results for $N \in \{1, 5, 10\}$. This label-based evaluation effectively measures the model's ability to maintain semantic and spatial consistency across diverse modalities such as text and event streams.

\begin{figure*}
\centering
\includegraphics[width=\textwidth]{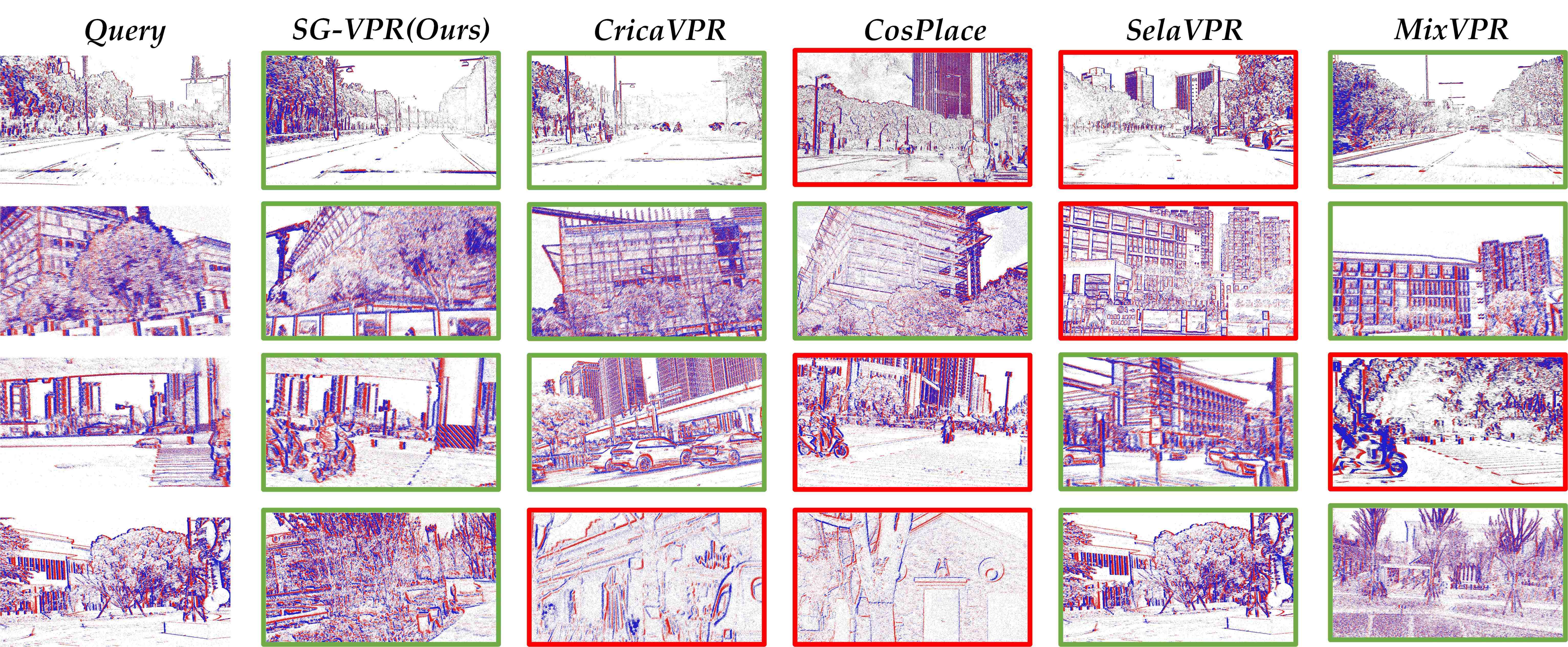}
\caption{Visualization of qualitative results. The leftmost column shows the queries, followed by the retrieval results from our SG-VPR and four baselines. Green/Red borders denote success/failure. Our method achieves correct retrieval across all samples, outperforming competitors that struggle with large viewpoint shifts and environmental noise.}
\label{fig:result_visualization}
\end{figure*}

\begin{figure*}
\centering
\includegraphics[width=\textwidth]{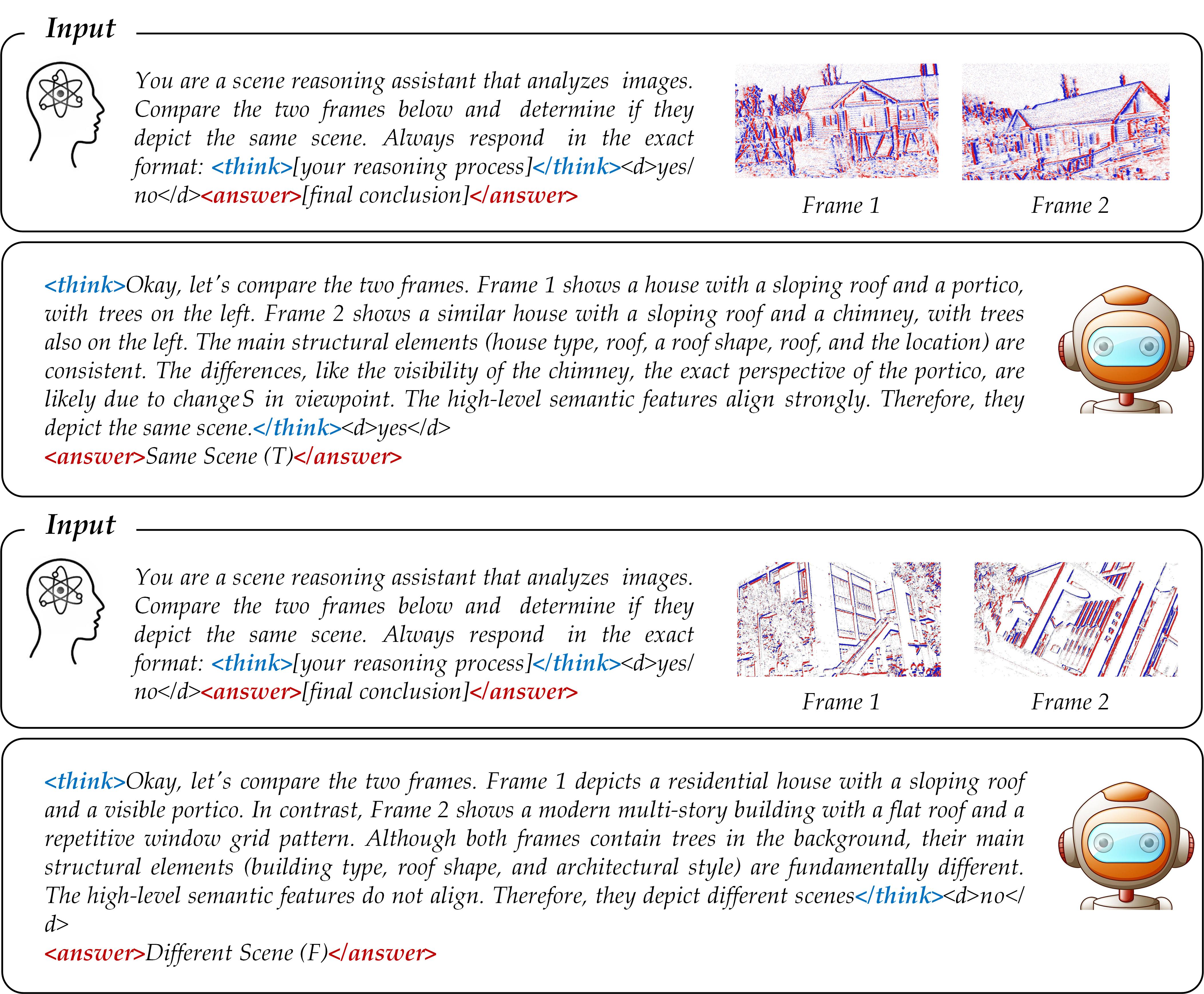}
\caption{\textbf{Visualization of the auxiliary reasoning process.} As shown in the ``think'' blocks, the model provides step-by-step semantic analysis to justify the retrieval results. By identifying shared landmarks (top) or structural mismatches (bottom), it offers human-understandable explanations for both successful matches and rejected candidates.} 
\label{fig:reasoning}
\end{figure*}

\subsection{Implementation Details}  
We implement our framework using PyTorch on a server equipped with a single NVIDIA RTX 3090 GPU. During the training phase, all input images are resized to $224 \times 224$ and normalized using ImageNet statistics. The model is optimized using the Adam optimizer with an initial learning rate of $1 \times 10^{-4}$ and a weight decay of $1 \times 10^{-3}$. To stabilize convergence, we employ a StepLR scheduler, decaying the learning rate by a factor of 0.5 every 3 epochs, with a batch size of 24. Following the efficient fine-tuning protocol of CricaVPR~\cite{cricavpr}, we freeze the parameters of the DINOv2~\cite{oquab2023dinov2} backbone and the CLIP~\cite{radford2021learning} Text Encoder to preserve their pre-trained generalized knowledge. Only the lightweight Adapter modules, the task-specific projection layers, and our proposed Cross-modal Fusion module are updated. This strategy allows for rapid domain adaptation without compromising the robust pre-trained feature space. More details can be found in our source code.

\subsection{Comparison with Other SOTA Algorithms} 
As shown in Table~\ref{tab:comparison-methods}, it reports the retrieval performance on two benchmark datasets. Our proposed SG-VPR achieves consistent improvements over existing state-of-the-art methods across most metrics.

$\bullet$ \textbf{Performance on EPRBench Dataset.~} 
We first evaluate the performance on our newly introduced EPRBench dataset to establish a baseline for future research. As shown in the rightmost column, our SG-VPR sets a strong state-of-the-art with 94.3\% R@1. It outperforms competitive methods like SuperVLAD~\cite{lu2024supervlad} (93.3\%) and CricaVPR~\cite{cricavpr} (92.6\%). This result not only validates the effectiveness of our semantic-guided design but also demonstrates the high quality and distinctiveness of the proposed benchmark.

$\bullet$ \textbf{Performance on NYC-Event-VPR Dataset.~} 
The advantages of our method are more pronounced on the challenging NYC-Event-VPR dataset (Event modality), which features complex urban environments and severe sparsity. While R2Former~\cite{xu2023r2former} and Pair-VPR~\cite{hausler2025pairvpr} achieve top performance using a two-stage retrieval method, our SG-VPR secures the second-best position with 57.5\% R@1, strictly outperforming specialized VPR methods like CricaVPR~\cite{cricavpr} (30.3\%) and MixVPR~\cite{ismail2022mixvpr} (45.7\%) by substantial margins of +27.2\% and +11.8\%, respectively. This result highlights that our method effectively mitigates the impact of noise and sparsity where traditional geometric-based methods fail.
Furthermore, on the NYC-Event-VPR (RGB modality) split, SG-VPR maintains competitive performance (86.5\% R@1), comparable to leading RGB-based methods, indicating the versatile generalization capability of our learned representations.

\begin{table}
    \centering
    \caption{{Ablation study of the proposed fusion modules.} Experiments are conducted on our self-collected \textbf{EPRBench} dataset. The best results are highlighted in bold.}
    \label{tab:ablation_fusion}
    \resizebox{\linewidth}{!}{
        \begin{tabular}{l|ccc}
            \toprule
            \textbf{Method} & \textbf{R@1} & \textbf{R@5} & \textbf{R@10} \\
            \midrule
            Baseline (Vision Only) & 92.6 & 95.3 & 96.5 \\
            \midrule
            + Global Fusion (\textit{cls}-fusion) & 93.7 & 95.4 & 96.8 \\
            + Local Fusion (\textit{patch}-fusion) & 93.8 & 95.8 & 96.8 \\
            \textbf{+ Full Fusion (Global + Local)} & \textbf{94.3} & \textbf{96.1} & \textbf{97.1} \\
            \bottomrule
        \end{tabular}
    }
\end{table}

\subsection{Ablation Study}

$\bullet$ \textbf{Analysis on Global and Local Semantics Fusion.} 
As shown in Table~\ref{tab:ablation_fusion}, we investigate the impact of semantic guidance at different scales. Introducing Global Fusion yields a 1.1\% gain in R@1, indicating that holistic context helps disambiguate visually similar scenes. Meanwhile, Local Fusion provides a 1.2\% improvement, proving its ability to mitigate dynamic interference in event streams by focusing on text-described landmarks. Most importantly, the Full Model achieves the best performance of 94.3\%. This confirms that global alignment and local sparsification are complementary, collectively ensuring robustness against the extreme conditions in the EPRBench dataset.

$\bullet$ \textbf{Analysis on Top-k Selection Strategy.} 
As reported in Table~\ref{tab:param_rho}, we investigate the sensitivity of the selection threshold $\rho$. Setting $\rho$ to a low value (e.g., 0.20) causes a significant performance drop to 93.1\%, indicating that an overly aggressive sparsification strategy results in the loss of critical structural cues. Conversely, increasing $\rho$ to 0.30 leads to a slight decline (94.2\%), suggesting that retaining excessive patches reintroduces task-irrelevant dynamic noise. The model achieves its peak performance of 94.3\% at $\rho=0.25$, confirming that this ratio strikes the optimal balance between filtering event artifacts and preserving semantic landmarks.

\begin{table}[t]
    \centering
    \caption{{Sensitivity analysis of the selection ratio $\rho$.} We evaluate the impact of different top-$k$ ratios on the EPRBench dataset. The default setting is $\rho=0.25$.}
    \label{tab:param_rho}
    \setlength{\tabcolsep}{4.5mm}
    \begin{tabular}{c|ccc}
        \toprule
        \textbf{Ratio $\rho$} & \textbf{R@1} & \textbf{R@5} & \textbf{R@10} \\
        \midrule
        0.15 & 93.9 & 95.9 & 96.6 \\
        0.20 & 93.1 & 95.6 & 96.7 \\
        \textbf{0.25} & \textbf{94.3} & \textbf{96.1} & \textbf{97.1} \\
        0.30 & 94.2 & 95.6 & 96.4 \\
        \bottomrule
    \end{tabular}
\end{table}

$\bullet$ \textbf{Analysis Study on Loss Weight $\gamma$.} 
The impact of the loss weight $\gamma$ is summarized in Table~\ref{tab:ablation_gamma}. By varying $\gamma$ from 0.10 to 0.25, we observe that the retrieval performance reaches its peak at $\gamma=0.15$ (94.3\% R@1). This suggests that incorporating a moderate auxiliary contrastive loss enhances feature discriminability. Conversely, further increasing $\gamma$ to 0.20 or 0.25 results in performance drops to 93.8\% and 93.6\%, respectively. We attribute this to the auxiliary loss overwhelming the primary retrieval supervision at higher weights. Based on these findings, we fix $\gamma=0.15$ for all subsequent experiments.

\begin{table}[h]
    \centering
    \caption{{Analysis study on the hyperparameter $\gamma$}. We investigate the impact of the weight $\gamma$ in the loss function (Eq.~\ref{e8}) on the EPRBench dataset. The best performance is achieved when $\gamma=0.15$.}
    \label{tab:ablation_gamma}
    \setlength{\tabcolsep}{3.5mm} 
    \begin{tabular}{c|ccc}
        \toprule
        \textbf{Hyperparameter} $\gamma$ & \textbf{R@1} & \textbf{R@5} & \textbf{R@10} \\
        \midrule
        0.10 & 92.9 & 95.7 & 96.9 \\
        \textbf{0.15} & \textbf{94.3} & \textbf{96.1} & \textbf{97.1} \\
        0.20 & 93.8 & 95.6 & 96.9 \\
        0.25 & 93.6 & 95.6 & 96.7 \\
        \bottomrule
    \end{tabular}
\end{table}

\subsection{Visualization} 

To provide an intuitive understanding of the proposed benchmark and our method's superiority, we present qualitative visualizations in Fig.~\ref{fig:datasets_visualization}, Fig.~\ref{fig:result_visualization}, and Fig.~\ref{fig:reasoning}. Fig.~\ref{fig:datasets_visualization} first highlights the diversity of EPRBench, which features significant viewpoint variations across different scene categories, creating a rigorous testing ground. Complementing this, Fig.~\ref{fig:result_visualization} confirms the effectiveness of our SG-VPR under these difficult conditions. As observed, while baseline methods frequently fail due to environmental noise or large viewpoint shifts, our model successfully leverages high-level textual priors to anchor retrieval on stable semantic landmarks, consistently achieving correct matches (green borders) where others fail.
Furthermore, Fig.~\ref{fig:reasoning} demonstrates the interpretability of our proposed Scene Expert. As shown in the visualization, the model employs a Chain-of-Thought (CoT) mechanism to perform a step-by-step comparison between the query and the retrieved candidate. Instead of relying solely on opaque feature distances, the system explicitly identifies and aligns shared high-level semantic structures to verify semantic consistency. This process not only validates the correctness of the match under significant viewpoint changes but also provides transparent, human-understandable justifications for the retrieval decision.

\subsection{Limitation Analysis}  

While the proposed EPRBench benchmark effectively mitigates the environmental constraints faced by RGB sensors (e.g., variations in lighting conditions), its reliance on a specifically fine-tuned scene expert model for text generation remains a core bottleneck in improving generalization capability. The currently used expert model is tailored and optimized for our specific data distribution, which greatly limits its direct transferability to open-world RGB scenarios.

\section{Conclusion} \label{sec::conclusion}
In this paper, in order to address the issue of lacking challenging benchmarks for Event-based VPR, we propose a large-scale Event-based VPR dataset named EPRBench, which includes 13k samples equipped with fine-grained scene descriptions. Complementing the data, we propose SG-VPR, which integrates high-level textual semantics into the visual retrieval pipeline. Through global alignment and local aggregation, our method effectively resolves the ambiguity caused by sparse event streams. Experimental results validate that SG-VPR achieves state-of-the-art performance and exceptional robustness against geometric deformations. Moving forward, we plan to extend our approach by exploring a General-Purpose Scene Expert, aiming to achieve unified scene understanding for broader cross-modal applications.

\section*{Acknowledgment}   
This work was supported by the National Natural Science Foundation of China under Grant U24A20342, 62102205. Anhui Provincial Natural Science Foundation-Outstanding Youth Project, 2408085Y032. The authors acknowledge the High-performance Computing Platform of Anhui University for providing computing resources.

\small{ 
\bibliographystyle{IEEEtran}
\bibliography{reference}
}

\end{document}